\documentclass[review]{elsarticle}
\usepackage{hyperref}
\usepackage{url}
\usepackage{amsmath}
\usepackage{amssymb}
\usepackage{graphicx}
\usepackage{wrapfig}
\usepackage{subfig}
\usepackage{float}
\usepackage{bbm}
\usepackage{CJKutf8}
\usepackage{footnote}
\usepackage{tablefootnote}
\DeclareMathOperator*{\argmax}{argmax}
\usepackage{hyperref}
\journal{Journal of \LaTeX\ Templates}

\begin{document}

\begin{frontmatter}

\title{On Extended Long Short-term Memory and Dependent
Bidirectional Recurrent Neural Network}

\author[USC]{Yuanhang Su\corref{cor1}}
\author[USC]{C.-C. Jay Kuo}
\cortext[cor1]{Corresponding Author\\ 
Email address: suyuanhang@hotmail.com}
\address[USC]{University of Southern California, Ming Hsieh
Department of Electrical Engineering, 3740 McClintock Avenue, Los
Angeles, CA, United States}

\begin{abstract}
In this work, we first analyze the memory behavior in three recurrent
neural networks (RNN) cells; namely, the simple RNN (SRN), the long
short-term memory (LSTM) and the gated recurrent unit (GRU), where the
memory is defined as a function that maps previous elements in a
sequence to the current output. Our study shows that all three of them
suffer rapid memory decay. Then, to alleviate this effect, we introduce
trainable scaling factors that act like an attention mechanism to adjust
memory decay adaptively. The new design is called the extended LSTM
(ELSTM).  Finally, to design a system that is robust to previous
erroneous predictions, we propose a dependent bidirectional recurrent
neural network (DBRNN). Extensive experiments are conducted on different
language tasks to demonstrate the superiority of the proposed ELSTM and
DBRNN solutions. The ELTSM has achieved up to 30\% increase in the
labeled attachment score (LAS) as compared to LSTM and GRU in the
dependency parsing (DP) task.  Our models also outperform other
state-of-the-art models such as bi-attention \cite{BiDecoderParsing} and
convolutional sequence to sequence (convseq2seq) \cite{ConvSeq2seq} by
close to 10\% in the LAS. The code is released as an open source\footnote{
https://github.com/yuanhangsu/ELSTM-DBRNN}.

\end{abstract}


\end{frontmatter}

\section{Introduction}\label{introduction}

We are interested in the design of an effective
sequence learning system to address sequence-in-sequence-out (SISO)
problems.  One key question lies in memory capability of the system --
how to retain sufficient input information in the learning process. In
this work, we examine the memory of a particular type of learning system
called the recurrent neural network (RNN). The recurrent neural network (RNN) has proved to be an effective
solution for natural language processing (NLP) through the advancement
in the last three decades \cite{Time,Jordan}. At the cell level, the
long short-term memory (LSTM) \cite{LSTM} and the gated recurrent unit
(GRU) \cite{GRU} are often adopted by an RNN as its low-level building
element.  Built upon these cells, various RNN models have been proposed. To name a few, there are the bidirectional RNN (BRNN) \cite{BRNN}, the encoder-decoder model \cite{GRU,Seq2Seq,Grammar,BiEnc_Dec} and the deep RNN \cite{Deep_RNN}. 

LSTM and GRU cells were designed to enhance the memory length of RNNs
and address the gradient vanishing/exploding issue
\cite{LSTM,OnDifficult, Difficult}, yet thorough analysis on their
memory decay property is lacking. The first objective of this research
is to analyze the memory length of three RNN cells - simple RNN (SRN)
\cite{Time,Jordan}, LSTM and GRU.  It will be conducted in Sec.
\ref{sec:memory}. Our analysis is different from the investigation of
gradient vanishing/exploding problem in the following sense. The
gradient vanishing/exploding problem occurs in the training process
while memory analysis is conducted on a trained RNN model.  Based on the
analysis, we further propose a new design in Sec.  \ref{sec:ELSTM} to
extend the memory length of a cell, and call it the extended long
short-term memory (ELSTM). 

As to the macro RNN model, one popular choice is the BRNN \cite{BRNN}.
Another choice is the encoder-decoder system, where the attention mechanism was introduced to improve its performance in \cite{Grammar,BiEnc_Dec}.  We show that the encoder-decoder system is not an efficient learner by itself. A better solution is to exploit the encoder-decoder and the BRNN jointly so as to overcome their individual limitations. Following this line of thought, we propose a new multi-task model, called the dependent bidirectional recurrent neural network (DBRNN), in Sec. \ref{sec:DBRNN}. 

To demonstrate the performance of the DBRNN model with the ELSTM cell,
we conduct a series of experiments on the language modeling (LM), the part of speech (POS) tagging
and the dependency parsing (DP) problems in Sec.  \ref{sec:experiment}.
Finally, concluding remarks are given and future research direction is
pointed out in Sec.  \ref{sec:conclusion}. 

\section{Memory Analysis of SRN, LSTM and GRU}\label{sec:memory}

For a large number of NLP tasks, we are concerned with finding semantic
patterns from input sequences. It was shown by Elman \cite{Time} that an
RNN builds an internal representation of semantic patterns. The memory
of a cell characterizes its ability to map input sequences of certain
length into such a representation. Here, we define the memory as a
function that maps elements of the input sequence to the current output.
Thus, the memory of an RNN is not only about whether an element can be
mapped into the current output but also how this mapping takes place.
It was reported by Gers {\em et al.} \cite{forget} that an SRN only
memorizes sequences of length between 3-5 units while an LSTM could
memorize sequences of length longer than 1000 units.  In this section,
we conduct memory analysis on SRN, LSTM and GRU cells. 

\subsection{Memory of SRN}\label{memory_SRN}

For ease of analysis, we begin with Elman's SRN model \cite{Time} with a
linear hidden-state activation function and a non-linear output
activation function since such a cell model is mathematically tractable
while its performance is equivalent to Jordan's model \cite{Jordan}. 

The SRN model can be described by the following two equations:
\begin{eqnarray}
\underline{\boldsymbol{c}}_t &=& \boldsymbol{W}_c \underline{\boldsymbol{c}}_{t-1} + \boldsymbol{W}_{in} \underline{\boldsymbol{X}}_t, \label{eq:SRN_ct} \\
\underline{\boldsymbol{h}}_t &=& f(\underline{\boldsymbol{c}}_t), \label{eq:SRN_ht}
\end{eqnarray}
where subscript $t$ is the time unit index, $\boldsymbol{W}_c \in \mathbb{R}^{N
\times N}$ is the weight matrix for hidden-state vector $\underline{\boldsymbol{c}}_{t-1} \in
\mathbb{R}^{N}$, $\boldsymbol{W}_{in}\in \mathbb{R}^{N \times M}$ is the weight
matrix of input vector $\underline{\boldsymbol{X}}_t\in \mathbb{R}^{M}$, $\underline{\boldsymbol{h}}_t \in \mathbb{R}^{N}$
in the output vector, and $f(\cdot)$ is an element-wise non-linear
activation function.  Usually, $f(\cdot)$ is a hyperbolic-tangent or a
sigmoid function.  We express matrices, vectors and scalars by bold-face italic,
bold-face italic with a straight line below and non-bold italic,
respectively (e.g. $\boldsymbol{m}, \underline{\boldsymbol{v}}, $ and
$s$). We omit the bias terms by
including them in the corresponding weight matrices. The multiplication 
between two equal-sized vectors in this paper is element-wise 
multiplication.
 
By induction, $\underline{\boldsymbol{c}}_t$ can be written as 
\begin{equation}
\underline{\boldsymbol{c}}_t = \boldsymbol{W}_c^t \underline{\boldsymbol{c}}_0 + \sum^t_{k=1} \boldsymbol{W}_c^{t-k} \boldsymbol{W}_{in} \underline{\boldsymbol{X}}_k, \label{eq:SRN_c0}
\end{equation}
where $\underline{\boldsymbol{c}}_0$ is the initial internal state of the SRN. Typically, we 
set $\underline{\boldsymbol{c}}_0=\underline{0}$. Then, Eq. (\ref{eq:SRN_c0}) becomes
\begin{equation}
\underline{\boldsymbol{c}}_t = \sum^t_{k=1} \boldsymbol{W}_c^{t-k} \boldsymbol{W}_{in} \underline{\boldsymbol{X}}_k. \label{eq:SRN_c}
\end{equation}

As shown in Eq. (\ref{eq:SRN_c}), SRN's output is a
function of all proceeding elements in the input sequence. The
dependency between the output and the input allows the SRN to retain the
semantic sequential patterns from the input. For the rest of this paper,
we call a system whose function introduces dependency between the output
and its proceeding elements in the input as {\it a system with memory}.

Athough the SRN is a system with memory, its memory length is limited.
Let $\lambda_{\max}$ be the largest singular value of
$\boldsymbol{W}_c$. Then, we have
\begin{equation}
|\boldsymbol{W}_c^{t-k} \boldsymbol{W}_{in} \underline{\boldsymbol{X}}_k| \leq ||\boldsymbol{W}_c||^{t-k} |\boldsymbol{W}_{in} \underline{\boldsymbol{X}}_k| = \sigma_{\max}(\boldsymbol{W}_c)^{t-k}|\boldsymbol{W}_{in} \underline{\boldsymbol{X}}_k|, k \leq t.
\label{eq:SRN_mem_decay}
\end{equation}
where $||\cdot||$ denotes matrix norm and $|\cdot|$ denotes vector norm,
both are $l^2$ norm. The $\sigma_{\max}(\cdot)$ denotes the largest
singular value of. The inequality is derived by the definition of matrix
norm. The equality is derived by the fact that the spectral norm ($l^2$
norm of a matrix) of a square matrix is equal to its largest singular
value. 

Here, we are only interested in the case of memory decay when
$\sigma_{\max}(\boldsymbol{W}_c) < 1$.  Since the contribution of $\underline{\boldsymbol{X}}_k$, $k<t$, to output
$\underline{\boldsymbol{h}}_t$ decays at least in form of $\sigma_{\max}(\boldsymbol{W}_c) ^{t-k}$, we conclude
that {\bf SRN's memory decays at least exponentially with its memory length $t-k$}. 

\subsection{Memory of LSTM}\label{memory_LSTM}

\begin{figure}[htbp]
\begin{center}
\includegraphics[width = \linewidth]{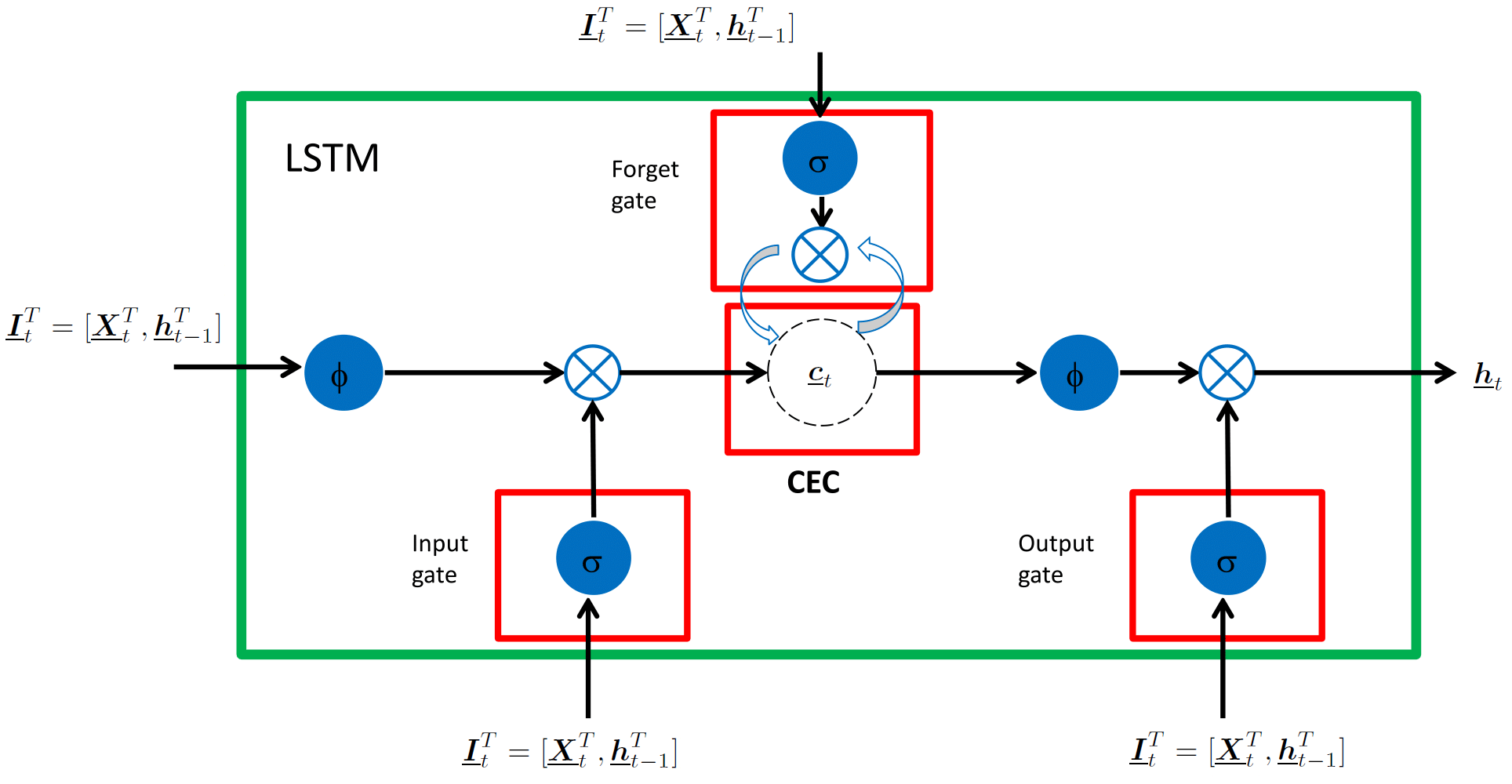}
\end{center}
\caption{The diagram of a LSTM cell.}\label{fig:LSTM}
\end{figure}

By following the work of Hochreiter {\em et al.} \cite{LSTM}, we plot
the diagram of the LSTM cell in Fig. \ref{fig:LSTM}. In this figure,
$\phi$, $\sigma$ and $\otimes$ denote the hyperbolic tangent function,
the sigmoid function (to be differed from the singular value operations denote as $\sigma_{\max}$ or $\sigma_{\min}$ with subscript) and the multiplication operation, respectively. All
of them operate in an element-wise fashion. The LSTM cell has an input
gate, an output gate, a forget gate and a constant error carousal (CEC)
module.  Mathematically, the LSTM cell can be written as
\begin{eqnarray}
\underline{\boldsymbol{c}}_t &=& \sigma (\boldsymbol{W}_f \underline{\boldsymbol{I}}_t) \underline{\boldsymbol{c}}_{t-1} + \sigma (\boldsymbol{W}_i \underline{\boldsymbol{I}}_t) \phi(\boldsymbol{W}_{in} \underline{\boldsymbol{I}}_t), 
\label{eq:LSTM_ct} \\
\underline{\boldsymbol{h}}_t &=& \sigma (\boldsymbol{W}_o \underline{\boldsymbol{I}}_t) \phi(\underline{\boldsymbol{c}}_t), \label{eq:LSTM_ht}
\end{eqnarray}
where $\underline{\boldsymbol{c}}_t \in \mathbb{R}^N$, column vector $\underline{\boldsymbol{I}}_t \in \mathbb{R}^{(M+N)}$
is a concatenation of the current input, $\underline{\boldsymbol{X}}_t \in \mathbb{R}^M$, and the
previous output, $\underline{\boldsymbol{h}}_{t-1} \in \mathbb{R}^N$ ({\em i.e.,} $\underline{\boldsymbol{I}}_t^T =
[\underline{\boldsymbol{X}}_t^T, \underline{\boldsymbol{h}}_{t-1}^T]$). Furthermore, $\boldsymbol{W}_f$, $\boldsymbol{W}_i$, $\boldsymbol{W}_o$ and $\boldsymbol{W}_{in}$ are
weight matrices for the forget gate, the input gate, the output gate and
the input, respectively.

Under the assumption $\underline{\boldsymbol{c}}_0 = \underline{0}$, the hidden-state vector 
of the LSTM can be derived by induction as
\begin{equation}\label{eq:LSTM_c}
\underline{\boldsymbol{c}}_t = \sum^t_{k=1} \underbrace{ \bigg[ \prod^t_{j=k+1} \sigma(\boldsymbol{W}_f \underline{\boldsymbol{I}}_j) 
\bigg] }_\text{forget gate} \sigma (\boldsymbol{W}_i \underline{\boldsymbol{I}}_k) \phi(\boldsymbol{W}_{in} \underline{\boldsymbol{I}}_k).
\end{equation}
By setting $f(\cdot)$ in Eq. (\ref{eq:SRN_ht}) to the hyperbolic-tangent
function, we can compare outputs of the SRN and the LSTM below:
\begin{eqnarray}
\underline{\boldsymbol{h}}_t^{SRN} & = & \phi \bigg(\sum^t_{k=1}\boldsymbol{W}_c^{t-k}\boldsymbol{W}_{in} \underline{\boldsymbol{X}}_k \bigg),
\label{eq:SRN_output} \\
\underline{\boldsymbol{h}}_t^{LSTM} & = & \sigma (\boldsymbol{W}_o \underline{\boldsymbol{I}}_t) \phi \bigg( \sum^t_{k=1} 
\underbrace{ \bigg[ \prod^t_{j=k+1} \sigma(\boldsymbol{W}_f \underline{\boldsymbol{I}}_j) \bigg]}_\text{forget gate} 
\sigma(\boldsymbol{W}_i \underline{\boldsymbol{I}}_k) \phi(\boldsymbol{W}_{in} \underline{\boldsymbol{I}}_k) \bigg). \label{eq:LSTM_output} 
\end{eqnarray}
We see from the above that $\boldsymbol{W}_c^{t-k}$ and $\prod^t_{j=k+1} \sigma(\boldsymbol{W}_f
\underline{\boldsymbol{I}}_j)$ play the same memory role for the SRN and the LSTM, respectively. 

We can find many special cases where LSTM memory length exceeds SRN regardless of the choice of SRN's model parameters ($\boldsymbol{W}_c$, $\boldsymbol{W}_{in}$). For example

$$
\exists \boldsymbol{W}_f \quad s.t. \quad \min |\sigma (\boldsymbol{W}_f \underline{\boldsymbol{I}}_j)| \geq \sigma_{\max}(\boldsymbol{W}_c), \quad \forall \sigma_{\max}(\boldsymbol{W}_c) \in 
[0,1),
$$ 
then
\begin{equation}\label{eq:forget_mem_decay}
\Bigg|\prod^t_{j=k+1}\sigma(\boldsymbol{W}_f \underline{\boldsymbol{I}}_j) \Bigg| \geq \sigma_{\max}(\boldsymbol{W}_c)^{t-k}, t \geq k.
\end{equation}
As given in Eqs. (\ref{eq:SRN_mem_decay}) and
(\ref{eq:forget_mem_decay}), the impact of input $\underline{\boldsymbol{I}}_k$ on the output of
the LSTM lasts longer than that of the SRN. This means {\bf there always exists a LSTM whose memory length is longer than SRN for all possible choices of SRN}.

Conversely, to find a SRN with similar advantage to LSTM, we need to make sure $||\boldsymbol{W}_c^{t-k}|| \geq 1 \geq \Bigg|\prod^t_{j=k+1}\sigma(\boldsymbol{W}_f \underline{\boldsymbol{I}}_j) \Bigg|$. Although such $\boldsymbol{W}_c$ exists, this condition would easily leads to memory explosion. For example, one close lower bound for $||\boldsymbol{W}_c^{t-k}||$ is $\sigma_{\min}(\boldsymbol{W}_c)^{t-k}$, where $\sigma_{\min}(\boldsymbol{W}_c)$ is the smallest singular value of $\boldsymbol{W}_c$ (this comes from the fact of $||\boldsymbol{AB}|| \geq \sigma_{min}(\boldsymbol{A})||\boldsymbol{B}||$ and $||\boldsymbol{B}|| = \sigma_{\max}(\boldsymbol{B}) \geq \sigma_{min}(\boldsymbol{B})$, use induction for derivation). We need $\sigma_{\min}(\boldsymbol{W}_c) \geq 1$, and since $||\boldsymbol{W}_c^{t-k}|| \geq \sigma_{\min}(\boldsymbol{W}_c)^{t-k}$, the SRN's memory will grow exponentially and end up in memory explosion. Such memory explosion constraint does not exist in LSTM.

\subsection{Memory of GRU}\label{memory_GRU}

\begin{figure}[htbp]
\begin{center}
\includegraphics[width = 0.9\linewidth]{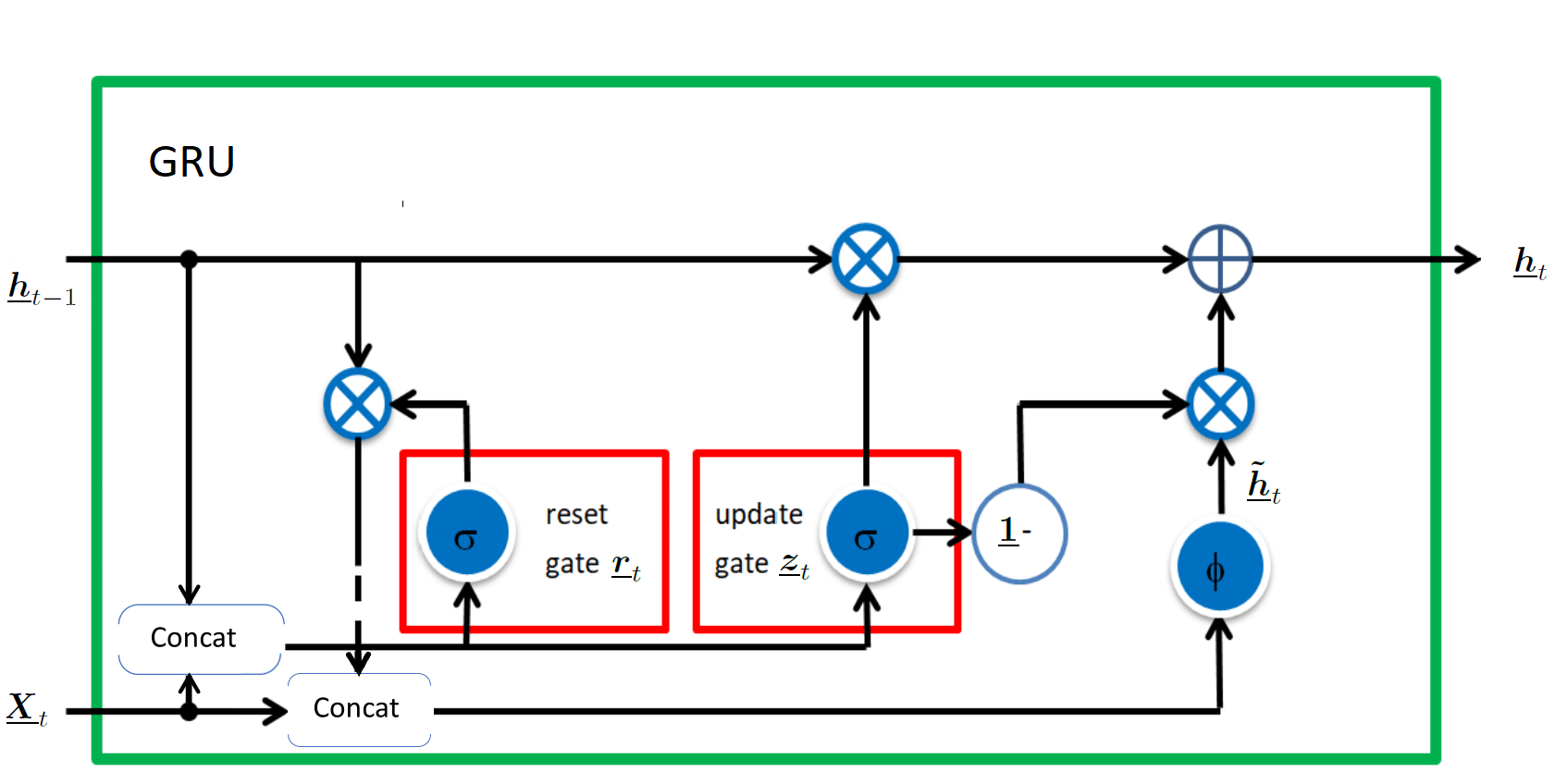}
\end{center}
\caption{The diagram of a GRU cell.}\label{fig:GRU}
\end{figure}

The GRU was originally proposed for neural machine translation
\cite{GRU}. It provides an effective alternative for the LSTM.  Its
operations can be expressed by the following four equations:
\begin{eqnarray}
\underline{\boldsymbol{z}}_t &=& \sigma (\boldsymbol{W}_z \underline{\boldsymbol{X}}_t + \boldsymbol{U}_z \underline{\boldsymbol{h}}_{t-1}), \label{eq:GRU_ut}\\
\underline{\boldsymbol{r}}_t &=& \sigma (\boldsymbol{W}_r \underline{\boldsymbol{X}}_t + \boldsymbol{U}_r \underline{\boldsymbol{h}}_{t-1}), \label{eq:GRU_rt}\\
\underline{\boldsymbol{\tilde h}}_t &=& \phi(\boldsymbol{W} \underline{\boldsymbol{X}}_t + \boldsymbol{U} (\underline{\boldsymbol{r}}_t \otimes \underline{\boldsymbol{h}}_{t-1})), \label{eq:GRU_tilde_ht} \\
\underline{\boldsymbol{h}}_t &=& \underline{\boldsymbol{z}}_t \underline{\boldsymbol{h}}_{t-1} + (\underline{\boldsymbol{1}}-\underline{\boldsymbol{z}}_t) \underline{\boldsymbol{\tilde h}}_t, \label{eq:GRU_ht}
\end{eqnarray}
where $\underline{\boldsymbol{X}}_t$, $\underline{\boldsymbol{h}}_t$, $\underline{\boldsymbol{z}}_t$ and $\underline{\boldsymbol{r}}_t$ denote the input, the hidden-state,
the update gate and the reset gate vectors, respectively, and $\boldsymbol{W}_z$,
$\boldsymbol{W}_r$, $\boldsymbol{W}$, are trainable weight matrices. Its hidden-state is also its
output, which is given in Eq.  (\ref{eq:GRU_ht}). Its diagram is shown in Fig. \ref{fig:GRU}, where Concat denotes the vector concatenation operation.

By setting $\boldsymbol{U}_z$, $\boldsymbol{U}_r$ and $\boldsymbol{U}$ to zero matrices, we can obtain the following simplified
GRU system:
\begin{eqnarray}
\underline{\boldsymbol{z}}_t &=& \sigma (\boldsymbol{W}_z \underline{\boldsymbol{X}}_t), \label{eq:GRU_new_ut}\\
\underline{\boldsymbol{\tilde h}}_t &=& \phi(\boldsymbol{W} \underline{\boldsymbol{X}}_t), \label{eq:GRU_new_tilde_ht} \\
\underline{\boldsymbol{h}}_t &=& \underline{\boldsymbol{z}}_t \underline{\boldsymbol{h}}_{t-1} + (\underline{\boldsymbol{1}}-\underline{\boldsymbol{z}}_t) \underline{\boldsymbol{\tilde h}}_t. \label{eq:GRU_new_ht}
\end{eqnarray}
For the simplified GRU with the initial rest condition, we can 
derive the following by induction:
\begin{equation} \label{eq:GRU_h}
\underline{\boldsymbol{h}}_t = \sum^t_{k=1} \bigg[ \underbrace{ \prod^t_{j=k+1} 
\sigma(\boldsymbol{W}_z \underline{\boldsymbol{X}}_j) }_\text{update gate} \bigg] (\underline{\boldsymbol{1}}-\sigma(\boldsymbol{W}_z \underline{\boldsymbol{X}}_k)) \phi(\boldsymbol{W} \underline{\boldsymbol{X}}_k).
\end{equation}

By comparing Eqs. (\ref{eq:LSTM_c}) and (\ref{eq:GRU_h}), we see that
the update gate of the simplified GRU and the forget gate of the LSTM
play the same role. In other words, {\bf there is no
fundamental difference between GRU and LSTM}. Such finding is
substantiated by the non-conclusive performance comparison between GRU
and LSTM conducted in \cite{LSTM_Odyssey,GRU_Empirical,Visualize_RNN}.

Because of the presence of the multiplication term
introduced by the forget gate and the update gate in Eqs.
(\ref{eq:LSTM_c}) and (\ref{eq:GRU_h}), the longer the distance of
$t-k$, the smaller these terms. Thus, the memory responses of LSTM and
GRU to $I_k$ diminish inevitably as $t-k$ becomes larger. This
phenomenon occurs regardless of the choice of model parameters.  For
complex language tasks that require long memory responses such as
sentence parsing, LSTM's and GRU's memory decay may have significant
impacts to their performance.

\section{Extended Long Short-Term Memory (ELSTM)}\label{sec:ELSTM}

\begin{figure}[htbp]
\begin{center}
\includegraphics[width = \linewidth]{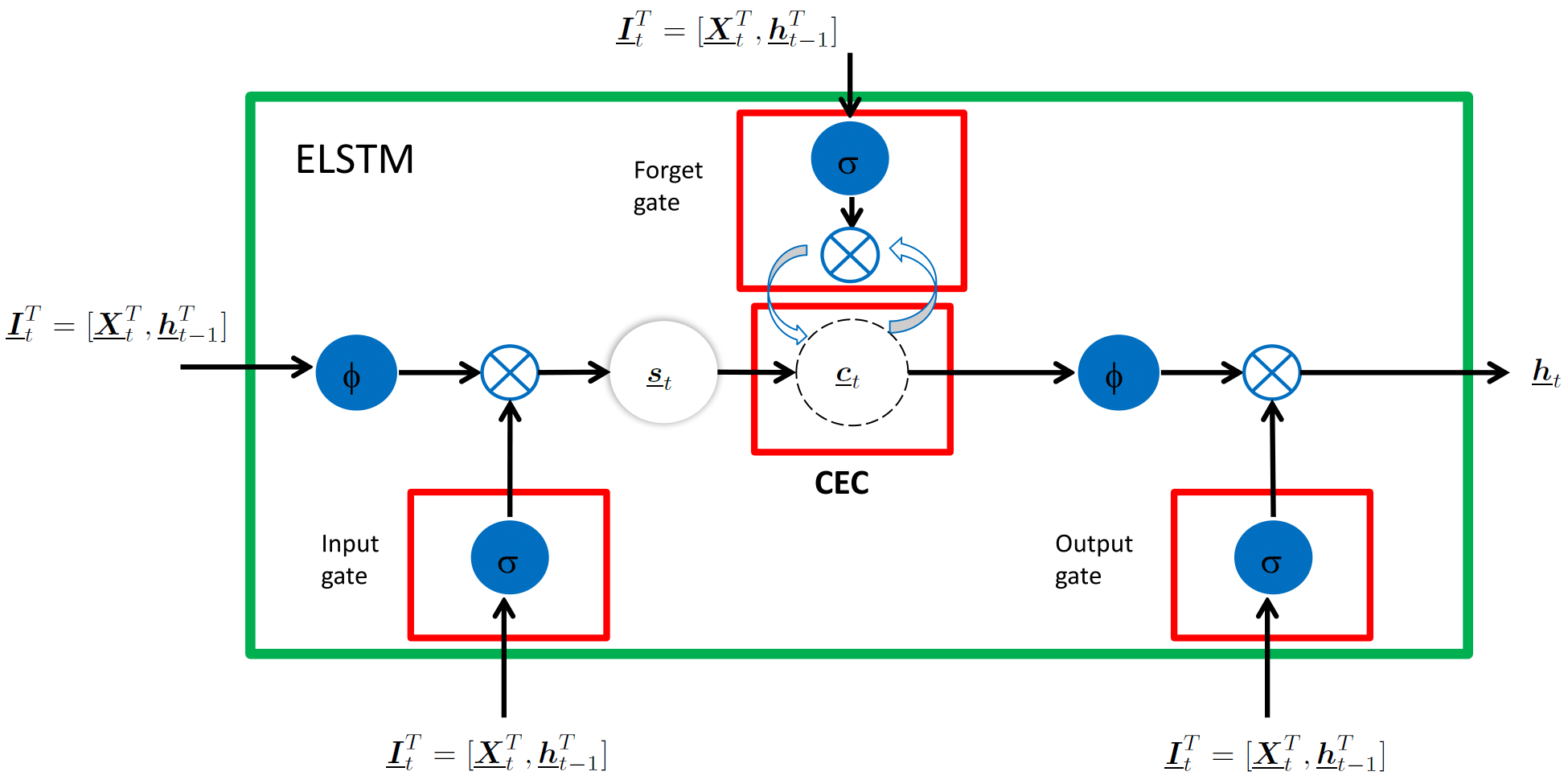}
\end{center}
\caption{The diagrams of the ELSTM cell.}\label{fig:ELSTM}
\end{figure}

To address this design limitation, we introduce a scaling
factor to compensate the fast decay of the input response. This leads to
a new solution called the extended LSTM (ELSTM). The ELSTM cell is
depicted in Fig. \ref{fig:ELSTM}, where $\underline{\boldsymbol{s}}_i
\in \mathbb{R}^N$, $i=1,\cdots,t-1$ is the trainable input scaling
vectors

The ELSTM cell can be described by
\begin{eqnarray}
\underline{\boldsymbol{c}}_t &=& \sigma (\boldsymbol{W}_f \underline{\boldsymbol{I}}_t) \underline{\boldsymbol{c}}_{t-1} + \underline{\boldsymbol{s}}_t \sigma (\boldsymbol{W}_i \underline{\boldsymbol{I}}_t)\phi(\boldsymbol{W}_{in} \underline{\boldsymbol{I}}_t), 
\label{eq:ELSTM_c_ct}\\
\underline{\boldsymbol{h}}_t &=& \sigma (\boldsymbol{W}_o \underline{\boldsymbol{I}}_t) \phi(\underline{\boldsymbol{c}}_t). \label{eq:ELSTM_c_ht}
\end{eqnarray}
a bias term $\underline{\boldsymbol{b}} \in \mathbb{R}^N$ for $\underline{\boldsymbol{c}}_t$ is omitted in Equation \ref{eq:ELSTM_c_ht}.
As shown above, we introduce scaling factor, $\underline{\boldsymbol{s}}_i$, $i= 1, \cdots,t-1$,
to the ELSTM to increase or decrease the impact of input $\underline{\boldsymbol{I}}_i$ in the sequence. 

To show that the ELSTM has longer memory than the LSTM, we first derive
a closed form expression of $\underline{\boldsymbol{h}}_t$ as
\begin{equation}
\underline{\boldsymbol{h}}_t = \sigma (\boldsymbol{W}_o \underline{\boldsymbol{I}}_t)\phi \bigg( \sum^t_{k=1} \underline{\boldsymbol{s}}_k \bigg[ \prod^t_{j=k+1} 
\sigma(\boldsymbol{W}_f \underline{\boldsymbol{I}}_j) \bigg] \sigma (\boldsymbol{W}_i \underline{\boldsymbol{I}}_k ) \phi(\boldsymbol{W}_{in} \underline{\boldsymbol{I}}_k )   \bigg). 
\label{eq:ELSTM_c_h}
\end{equation}
Then, we can find the following special case:
\begin{equation} \label{eq:ELSTM_vs_LSTM}
\exists \underline{\boldsymbol{s}}_k \quad s.t.\quad \Bigg|\underline{\boldsymbol{s}}_k \prod^t_{j=k+1}\sigma(\boldsymbol{W}_f \underline{\boldsymbol{I}}_j) \Bigg| \geq 
\Bigg|\prod^t_{j=k+1}\sigma(\boldsymbol{W}_f \underline{\boldsymbol{I}}_j) \Bigg| \quad \forall \boldsymbol{W}_f.
\end{equation}

By comparing Eq. (\ref{eq:ELSTM_vs_LSTM}) with Eq.
(\ref{eq:forget_mem_decay}), we conclude that {\bf there always exists an ELSTM whose memory is longer than LSTM for all choices of LSTM}. Conversely, we cannot find such LSTM with similar advantage to ELSTM. This demonstrates the ELSTM's system advantage by design to LSTM. Such a scalar-based solution can also be found in
work for convolutional neural networks (CNN) \cite{SENet}.

The numbers of parameters used by various RNN cells
are compared in Table \ref{tab:param_comp}, where $\underline{\boldsymbol{X}}_t \in
\mathbb{R}^M$, $\underline{\boldsymbol{h}}_t \in \mathbb{R}^N$ and $t = 1, \cdots, T$. As shown in Table \ref{tab:param_comp}, the number of parameters of the
ELSTM cell depends on the maximum length, $T$, of the input sequences,
which makes the model size uncontrollable. To address this problem, we
choose a fixed $T_s$ (with $T_s < T$) as the upper bound on the number of
scaling factors, and set $\underline{\boldsymbol{s}}_k = \underline{\boldsymbol{s}}_{(k-1) \mbox{ mod } T_s+1}$, if $k >
T_s$ and $k$ starts from 1, where mod denotes the modulo operator.  In
other words, the sequence of scaling factors is a periodic one with
period $T_s$, so the elements in a sequence that are distanced by the
length of $T_s$ will share the same scaling factor. 

\begin{table}[htbp]
\centering
\caption{Comparison of Parameter Numbers.}\label{tab:param_comp}
\begin{tabular*}{0.6\textwidth}{@{\extracolsep{\fill} }l c }
\multicolumn{1}{c}{\bf Cell} & \multicolumn{1}{c}{\bf Number of Parameters} \\ 
\hline
LSTM & $4N(M + N + 1)$\\
GRU & $3N(M + N + 1)$ \\
ELSTM & $4N(M + N + 1) + N (T+1)$ \\
\end{tabular*}
\end{table}

The ELSTM cell with periodic scaling factors can be described by
\begin{eqnarray}
\underline{\boldsymbol{c}}_t &=& \sigma (\boldsymbol{W}_f \underline{\boldsymbol{I}}_t) \underline{\boldsymbol{c}}_{t-1} + \underline{\boldsymbol{s}}_{t_s} \sigma (\boldsymbol{W}_i \underline{\boldsymbol{I}}_t)\phi(\boldsymbol{W}_{in} \underline{\boldsymbol{I}}_t), 
\label{eq:ELSTM_shared_c_ct}\\
\underline{\boldsymbol{h}}_t &=& \sigma (\boldsymbol{W}_o \underline{\boldsymbol{I}}_t) \phi(\underline{\boldsymbol{c}}_t), \label{eq:ELSTM_shared_c_ht}
\end{eqnarray}
where $t_s = (t-1) \mbox{ mod } T_s + 1$. 
We observe that the choice of $T_s$ affects the network performance.
Generally speaking, a small $T_s$ value is suitable for simple language
tasks that demand shorter memory while a larger $T_s$ value is desired
for complex ones that demand longer memory. For the particular
sequence-to-sequence (seq2seq \cite{Seq2Seq,Grammar}) RNN models, a
larger $T_s$ value is always preferred.  We will elaborate the parameter
settings in Sec. \ref{sec:experiment}. 

\subsection{Study of Scaling Factor}\label{sec:ELSTM_SF}

\begin{figure}[htbp]
\begin{center}
\includegraphics[width = 0.6\linewidth]{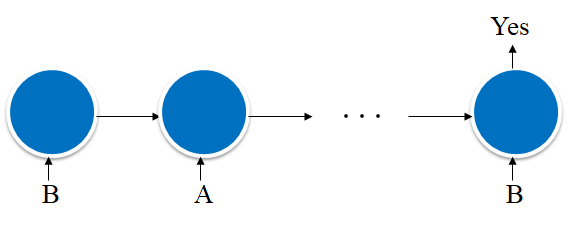}
\end{center}
\caption{Experiment of estimating the presence of ``A".}\label{fig:ELSTM_toy}
\end{figure}

To examine the memory capability of the scaling factor, we carry out the
following experiment. The RNN cell is asked to tell whether a special
element ``A" exists in the sequence of a single ``A" and multiple ``B"s
of length $T$. The training data contains $T$ number of positive samples
where ``A" locates from position 1 to $T$, and 1 negative sample where
there is no ``A" exists. The cell takes in the whole sequence and
generates the output at time step $T$ as shown in Fig.
\ref{fig:ELSTM_toy}. 

We would like to see the memory response of LSTM and ELSTM to ``A". If ``A" lies at the beginning of the sequence, the LSTM's memory decay may cause it lose the information of ``A"'s presence. The memory responses of LSTM and ELSTM to the input $\underline{\boldsymbol{I}}_k$ are calculated as:

\begin{eqnarray}
\underline{\boldsymbol{mr}}^{LSTM}_k &=& \bigg[ \prod^T_{j=k+1}  \sigma(\boldsymbol{W}_f \underline{\boldsymbol{I}}_j) \bigg] \sigma (\boldsymbol{W}_i \underline{\boldsymbol{I}}_k ) \phi(\boldsymbol{W}_{in} \underline{\boldsymbol{I}}_k ) , 
\label{eq:LSTM_ms}\\
\underline{\boldsymbol{mr}}^{ELSTM}_k &=& \underline{\boldsymbol{s}}_k \bigg[ \prod^T_{j=k+1}  \sigma(\boldsymbol{W}_f \underline{\boldsymbol{I}}_j) \bigg] \sigma (\boldsymbol{W}_i \underline{\boldsymbol{I}}_k ) \phi(\boldsymbol{W}_{in} \underline{\boldsymbol{I}}_k ) , 
\label{eq:ELSTM_ms}
\end{eqnarray}

The detailed model settings can be found in Table. \ref{tab:ELSTM_toy}

\begin{table}[htbp]
\centering
\caption{Network parameters for the toy experiment.}\label{tab:ELSTM_toy}
\begin{tabular*}{0.5\textwidth}{@{\extracolsep{\fill} }l c } \hline
Number of RNN layers& 1 \\
Embedding layer vector size& 2\\ 
Number of RNN cells & 1\\
Batch size & 5  \\ 
\end{tabular*}
\end{table}

We carry out multiple such experiments by increase the sample length $T$ by 1 at a time and see when LSTM cannot keep up with ELSTM. We train the LSTM and ELSTM models with equal number of epochs until both report no further change of training loss.

\begin{figure}[htbp]
\centering
\subfloat[]{\label{fig:comp_toy_short} \includegraphics[width = 
0.5\linewidth]{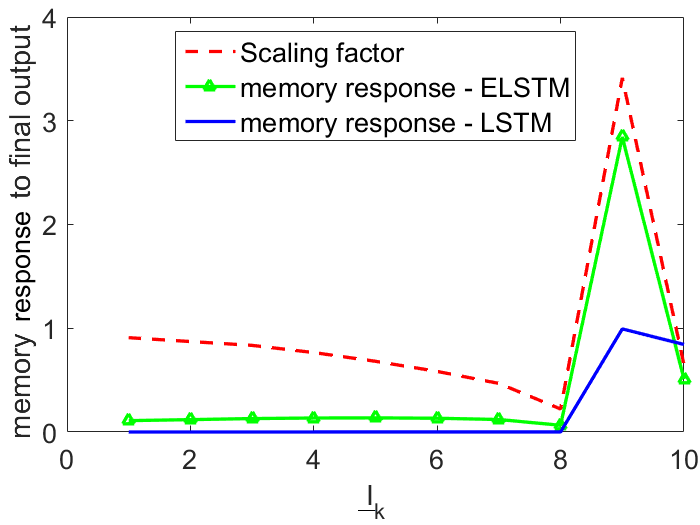}} 
\centering
\subfloat[]{\label{fig:comp_toy_long} \includegraphics[width = 
0.5\linewidth]{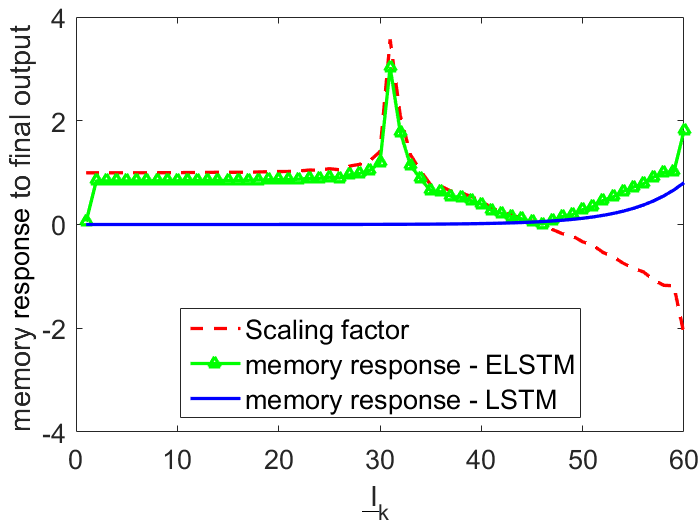}}
\caption{Comparison of memory response between LSTM and ELSTM.}\label{fig:comp_toy}
\end{figure}

We found when $T=60$, LSTM's training loss starts to plateau while ELSTM can further decrease to zero. As a result, LSTM starts to ``forget" when $T>=60$. The detailed plot of the memory responses for two particular samples are shown in Fig. \ref{fig:comp_toy}

Fig. \ref{fig:comp_toy_short} shows the memory response of trained LSTM and ELSTM on a sample with $T=10$ with ``A" at position 9. It can be seen that although both LSTM and ELSTM have stronger memory response at ``A", the ELSTM attends better than LSTM since its response at position 10 is smaller than LSTM's. We can also find that the scaling factor has larger value at the beginning and then slowly decreases as the location comes closer to the end. It then spikes at position 9. We can imagine that the scaling factor is doing its compensating job at both ends of the sequence.

Fig. \ref{fig:comp_toy_long} shows the memory response of trained LSTM and ELSTM on a sample with $T = 60$ with ``A" at location 30. In this case, the LSTM is not able to ``remember" the presence of ``A" and it does not have strong response to it. The scaling factor is doing its compensating job at the first half the sequence and especially in the middle and this causes strong ELSTM's response to ``A".

Even though scaling factor cannot adaptively change its value once it is trained, it is able to learn the pattern of model's rate of memory decay and the averaged importance of that position in the training set.

It is important to point out that the scaling factor needs to be initialized to 1 for each cell.

\section{Dependent BRNN (DBRNN) Model}\label{sec:DBRNN}

A single RNN cell is rarely used in practice due to its
limited capability in modeling real-world problems. To build a more
powerful RNN model, it is often to integrate several cells with
different probabilistic models. To give an example, the
sequence-to-sequence problem demands an RNN model to predict an output
sequence, $\{\underline{\boldsymbol{Y}}_t\}^{T'}_{t=1}$ with
$\underline{\boldsymbol{Y}}_i \in \mathbb{R}^N$, based on an input
sequence, $\{\underline{\boldsymbol{X}}_t\}^{T}_{t=1}$ with
$\underline{\boldsymbol{X}}_i \in \mathbb{R}^M$, where $T$ and $T'$ are
lengths of the input and the output sequences, respectively. To solve
this problem, we propose a macro RNN model, called the dependent BRNN
(DBRNN), in this section.  Our design is inspired by pros and cons of
two RNN models; namely, the bidirectional RNN (BRNN) \cite{BRNN} and the
encoder-decoder design \cite{GRU}. We will review the BRNN and the
encoder-decoder in Sec.  \ref{sec:DBRNN_BRNN_ENC_DEC}. Then, the DBRNN
is proposed in Sec.  \ref{DBRNN_model}. 

\subsection{BRNN and Encoder-Decoder}\label{sec:DBRNN_BRNN_ENC_DEC}

\begin{figure}[htbp]
\begin{center}
\includegraphics[width = 0.4\linewidth]{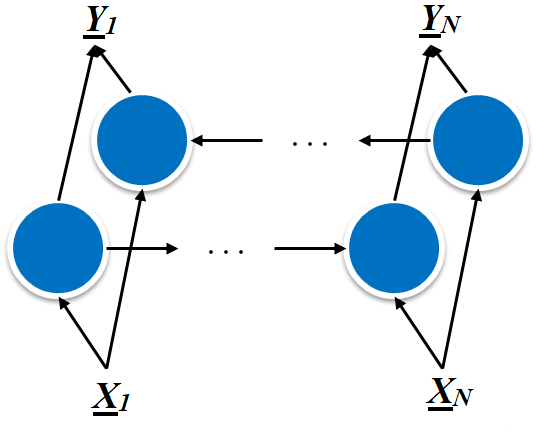}
\end{center}
\caption{The diagram of BRNN.}\label{fig:BRNN}
\end{figure}

As its name indicates, {\bf BRNN} takes inputs in both forward and
backward directions as shown in Fig. \ref{fig:BRNN}, and it has two RNN
cells to take in the input: one takes the input in the forward
direction, the other takes the input in the backward direction. 

The motivation for BRNN is to fully utilize the input sequence if future information ($\{\underline{\boldsymbol{X}}_i\}^T_{i=t+1}$) is accessible. This is especially helpful if current output $\underline{\boldsymbol{Y}}_t$ is also a function of future inputs. The conditional probability density function of BRNN is in form of

\begin{eqnarray} \label{eq:BRNN_model}
\underline{\boldsymbol{p}}_t &=& P(\underline{\boldsymbol{Y}}_t|\{\underline{\boldsymbol{X}}_i\}^{T}_{i=1}) = \boldsymbol{W}^f \underline{\boldsymbol{p}}^f_t + \boldsymbol{W}^b \underline{\boldsymbol{p}}^b_t, \label{eq:BRNN_model_pt} \\
\underline{\boldsymbol{\hat{Y}}}_t &=& \argmax_{\underline{\boldsymbol{Y}}_t} \underline{\boldsymbol{p}}_t,\label{eq:BRNN_model_Yt}
\end{eqnarray}

where

\begin{eqnarray}
\underline{\boldsymbol{p}}^f_t &=& P(\underline{\boldsymbol{Y}}_t|\{\underline{\boldsymbol{X}}_i\}^t_{i=1}) ,\\
\underline{\boldsymbol{p}}^b_t &=& P(\underline{\boldsymbol{Y}}_t|\{\underline{\boldsymbol{X}}_i\}^T_{i=t}) ,
\end{eqnarray}

and $\boldsymbol{W}^f$ and $\boldsymbol{W}^b$ are trainable weights, $\underline{\boldsymbol{\hat{Y}}}_t$ is the predicted output element at time step $t$. So the output is a combination of the density estimation of a forward RNN and the output of a backward RNN. Due to the bidirectional design, the BRNN can utilize the information of the entire input sequence to predict each individual output element. One example where such treatment is helpful is generating a sentence like ``this is an apple" for language modeling (predicts the next word given proceeding words in a sentence). In this case, the word ``an" strongly associates with its following word ``apple", in a forward directional RNN model, it would find difficulty in generating ``an" before ``apple".

\begin{figure}[htbp]
\begin{center}
\includegraphics[width = 0.4\linewidth]{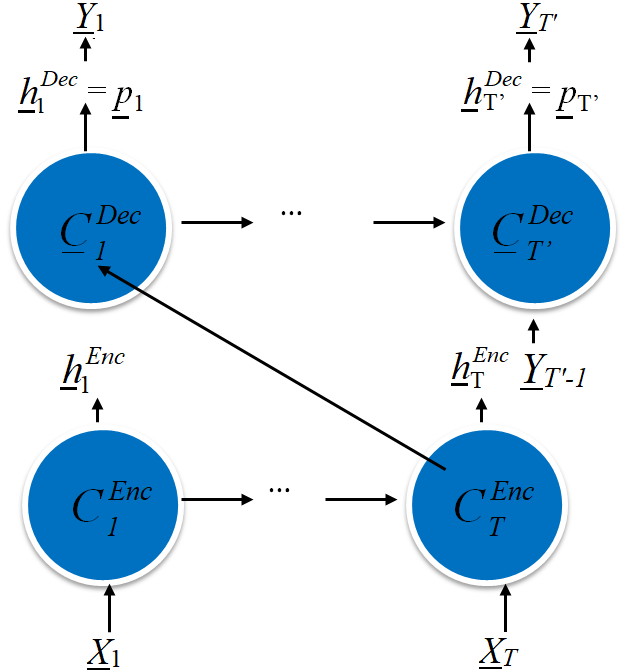}
\end{center}
\caption{The diagram of sequence to sequence (seq2seq).}\label{fig:ENC_DEC}
\end{figure}

{\bf Encoder-decoder} was first proposed for machine translation (MT) along with GRU in \cite{GRU}. It was motivated to handle the situation when $T' \neq T$. It has two RNN cells: an encoder and a decoder. The detailed design of one of the early proposals \cite{Seq2Seq} of encoder-decoder RNN model is illustrated in Fig. \ref{fig:ENC_DEC}.

As can be seen in Fig. \ref{fig:ENC_DEC}, the encoder (denoted by Enc) takes the input sequence of length $T$ and generates its output $\underline{\boldsymbol{h}}^{\text{Enc}}_i$ and hidden state $\underline{\boldsymbol{c}}^{\text{Enc}}_i$, where $i \in \{1,...,T\}$. In seq2seq model, the encoder's hidden state at time step $T$ is used as the representation of the input sequence. The decoder then utilizes the hidden state information to generate the output sequence of length $T'$ by initializing its hidden state $\underline{\boldsymbol{c}}^{Dec}_1$ as $\underline{\boldsymbol{c}}^{Enc}_T$. So the decoding process starts after the encoder has processed the entire input sequence. In practice, the input to the decoder at time step 1 is a pre-defined start decoding symbol. At the following time steps, the previous output $\underline{\boldsymbol{Y}}_{t-1}$ will be used as input. The decoder will stop the decoding process if a special pre-defined stopping symbol is generated.

As compare with BRNN, the encoder-decoder is not only advantageous in
its ability in handling input/output sequences of different length but
also capable in generating better aligned output sequences by explicitly
feeding previous predicted outputs back to its decoder. Thus, the
encoder-decoder estimates the following density function

\begin{eqnarray} \label{eq:ENC_DEC_model}
\underline{\boldsymbol{p}}_t &=& P(\underline{\boldsymbol{Y}}_t|\{\underline{\boldsymbol{\hat{Y}}}_i\}^{t-1}_{i=1},\{\underline{\boldsymbol{X}}_i\}^T_{i=1}) \\
\underline{\boldsymbol{\hat{Y}}}_t &=& \argmax_{\underline{\boldsymbol{Y}}_t} \underline{\boldsymbol{p}}_t \quad \forall t \in \{1,...,T'\}.
\end{eqnarray}

For better encoder-decoder aliment, various attention mechanism has been proposed for encoder-decoder model. In \cite{Grammar,BiEnc_Dec}, additional weighted connections are introduced to connect the decoder to the hidden state of the encoder.

On the other hand, the encoder-decoder system is vulnerable to previous
erroneous predictions in the forward path.  Recently, the BRNN was
introduced to the encoder by Bahdanau {\em et al.} \cite{BiEnc_Dec}, yet
their design does not address the erroneous prediction problem. 

\subsection{DBRNN Model and Training}\label{DBRNN_model}

Being motivated by the observations in Sec. \ref{sec:DBRNN_BRNN_ENC_DEC}, we
propose a multi-task BRNN model, called the dependent BRNN (DBRNN), to 
achieve the following objectives:
\begin{eqnarray}
\underline{\boldsymbol{p}}_t &=& \boldsymbol{W}^f \underline{\boldsymbol{p}}_t^f + \boldsymbol{W}^b \underline{\boldsymbol{p}}_t^b \label{eq:DBRNN_model_pt} \\
\underline{\boldsymbol{\hat{Y}}}^f_t &=& \argmax_{\underline{\boldsymbol{Y}}_t} \underline{\boldsymbol{p}}_t^f, \label{eq:DBRNN_model_Yft}\\
\underline{\boldsymbol{\hat{Y}}}^b_t &=& \argmax_{\underline{\boldsymbol{Y}}_t} \underline{\boldsymbol{p}}_t^b , \label{eq:DBRNN_model_Ybt} \\
\underline{\boldsymbol{\hat{Y}}}_t &=& \argmax_{\underline{\boldsymbol{Y}}_t} \underline{\boldsymbol{p}}_t \label{eq:DBRNN_model_Yt} 
\end{eqnarray}
where 
\begin{eqnarray}
\underline{\boldsymbol{p}}_t^f & = & P(\underline{\boldsymbol{Y}}_t|\{\underline{\boldsymbol{X}}_i\}^T_{i=1}, \{\underline{\boldsymbol{\hat{Y}}}^f_i\}^{t-1}_{i=1}), \\
\underline{\boldsymbol{p}}_t^b & = & P(\underline{\boldsymbol{Y}}_t|\{\underline{\boldsymbol{X}}_i\}^T_{i=1}, \{\underline{\boldsymbol{\hat{Y}}}^b_i\}^{T'}_{i=t+1}), \\
\underline{\boldsymbol{p}}_t   & = & P(\underline{\boldsymbol{Y}}_t|\{\underline{\boldsymbol{X}}_i\}^T_{i=1}), 
\end{eqnarray}
and $\boldsymbol{W}^f$ and $\boldsymbol{W}^b$ are trainable weights. As shown in Eqs.
(\ref{eq:DBRNN_model_Yft}), (\ref{eq:DBRNN_model_Ybt}) and
(\ref{eq:DBRNN_model_Yt}), the DBRNN has three learning objectives: 1)
the target sequence for the forward RNN prediction, 2) the reversed
target sequence for the backward RNN prediction, and 3) the target
sequence for the bidirectional prediction. 

\begin{figure}[htbp]
\centering
\includegraphics[width=0.4\linewidth]{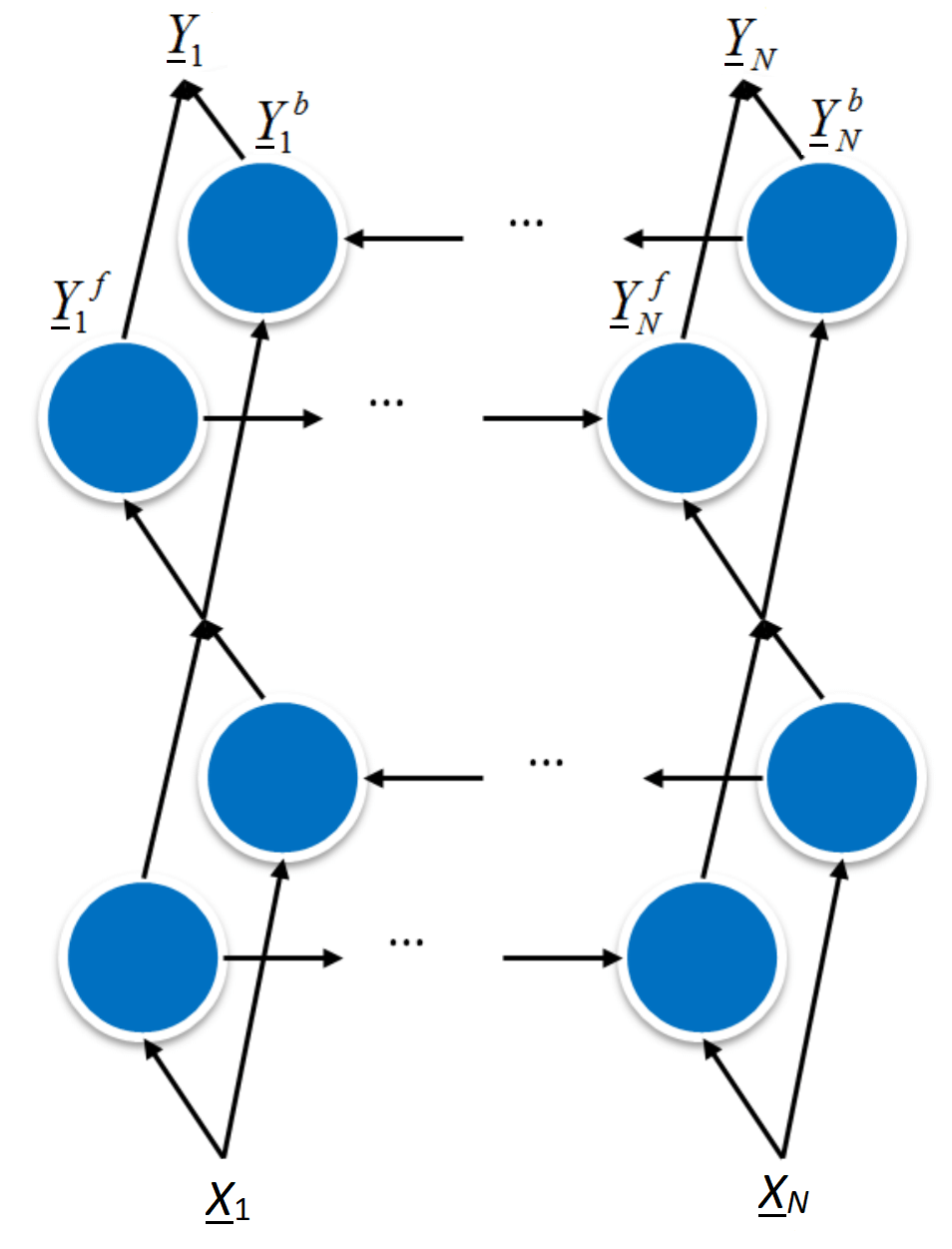}
\caption{The DBRNN model.}\label{fig:DBRNN}
\end{figure}

The DBRNN model is shown in Fig. \ref{fig:DBRNN}. It consists of a lower
and an upper BRNN branches. At each time step, the input to the forward
and the backward parts of the upper BRNN is the concatenated forward and
backward outputs from the lower BRNN branch. The final bidirectional
prediction is the pooling of both the forward and the backward
predictions.  We will show later that this design will make the DBRNN
robust to previous erroneous predictions. 

Let $F(\cdot)$ be the cell function.  The input is fed into the forward
and backward RNN of the lower BRNN branch as
\begin{equation} \label{eq:DBRNN_lBRNN}
\underline{\boldsymbol{h}}_t^f = F_l^f \Big(\underline{\boldsymbol{x}}_t, \underline{\boldsymbol{c}}_{l(t-1)}^f \Big), \quad
\underline{\boldsymbol{h}}_t^b = F_l^b \Big(\underline{\boldsymbol{x}}_t, \underline{\boldsymbol{c}}_{l(t+1)}^b \Big), \quad
\underline{\boldsymbol{h}}_t = \begin{bmatrix} \underline{\boldsymbol{h}}_t^f\\ \underline{\boldsymbol{h}}_t^b \end{bmatrix},
\end{equation}
where $\underline{\boldsymbol{c}}$ and $l$ denote the cell hidden state and the lower BRNN,
respectively.  The final output, $\underline{\boldsymbol{h}}_t$, of the lower BRNN is the
concatenation of the output, $\underline{\boldsymbol{h}}_t^f$, of the forward RNN and the output,
$\underline{\boldsymbol{h}}_t^b$, of the backward RNN. Similarly, the upper BRNN generates the
final output $\underline{\boldsymbol{p}}_t$ as
\begin{equation} \label{eq:DBRNN_uBRNN}
\underline{\boldsymbol{p}}_t^f = F_u^f \Big(\underline{\boldsymbol{h}}_t, \underline{\boldsymbol{c}}_{u(t-1)}^f \Big), \quad
\underline{\boldsymbol{p}}_t^b = F_u^b \Big(\underline{\boldsymbol{h}}_t, \underline{\boldsymbol{c}}_{u(t+1)}^b \Big), \quad
\underline{\boldsymbol{p}}_t = \boldsymbol{W}^f \underline{\boldsymbol{p}}_t^f + \boldsymbol{W}^b \underline{\boldsymbol{p}}_t^b,
\end{equation}
where $u$ denotes the upper BRNN. To generate forward prediction
$\underline{\boldsymbol{\hat{Y}}}^f_t$ and backward prediction $\underline{\boldsymbol{\hat{Y}}}^b_t$, the forward and
backward paths of the upper BRNN branch are separately trained by the
original and the reversed target sequences, respectively. The results of
forward and backward predictions of the upper RNN branch are then
combined to generate the final result. 

There are three errors: 1) forward prediction error $e_f$ for
$\underline{\boldsymbol{\hat{Y}}}^f_t$, 2) backward prediction error $e_b$ for $\underline{\boldsymbol{\hat{Y}}}^b_t$, and
3) bidirectional prediction error $e$ for $\underline{\boldsymbol{\hat{Y}}}_t$ .  To train the
proposed DBRNN, $e_f$ is backpropagated through time to the upper
forward RNN and the lower BRNN, $e_b$ is backpropagated through time to
the upper backward RNN and the lower BRNN, and $e$ is backpropagated
through time to the entire model. 

As it can been seen that DBRNN being an encoder-decoder can better handle output alignment. By introducing the bidirectional design to its decoder, DBRNN is also better than encoder-decoder in handling previous erroneous predictions. To show that DBRNN is more robust to previous erroneous predictions than 
one-directional models, we compare their cross entropy defined as
\begin{equation}
l = -\sum^K_{k=1} p_{t,k} \text{log}(\hat{p}_{t,k}),
\end{equation}
where $K$ is the total number of classes (e.g. the size of vocabulary
for the language task), $\hat{p}_t$ is the predicted distribution, 
and $p_t$ is the ground truth distribution with $k'$ as the ground truth 
label. It is in form of one-hot vector. That is,
$$
\underline{\boldsymbol{p}}_t = (\delta_{1,k'}, \cdots, \delta_{k',k'} , \cdots, \delta_{K,k'})^T, 
\quad k = 1, \cdots ,K,
$$ 
where $\delta_{k,k'}$ is the Kronecker delta function. Based on Eq. 
(\ref{eq:DBRNN_model_pt}), $l$ can be further expressed as
\begin{eqnarray}
l &=& -\sum^K_{k=1} p_{t,k} \text{log}(W^f_k \hat{p}^f_{t,k} + W^b_k \hat{p}^b_{t,k}), \\
  &=& -\text{log}(W^f_{k'} \hat{p}^f_{t,k'} + W^b_{k'} \hat{p}^b_{t,k'}).
\label{eq:DBRNN_cross_entropy}
\end{eqnarray}
We can select $W^f_{k'}$ and $W^b_{k'}$ such that $W^f_{k'} \hat{p}^f_{t,k'} + 
W^b_{k'} \hat{p}^b_{t,k'}$ is greater than  $\hat{p}^f_{t,k'}$ and 
$\hat{p}^b_{t,k'}$. Then, we obtain 
\begin{eqnarray}
l & < & -\sum^K_{k=1}\text{log}(\hat{p}^f_{tk}), \\
l & < & -\sum^K_{k=1}\text{log}(\hat{p}^b_{tk}).
\end{eqnarray}
The above two equations indicate that {\bf there always exists a DBRNN with better performance as compared to encoder-decoder regardless of which parameters the encoder-decoder chose}. So DBRNN does not have the encoder-decoder's model limitations.

It is worthwhile to compare the proposed DBRNN and the bi-attention model 
in Cheng {\em et al.} \cite{BiDecoderParsing}. Both of them have bidirectional 
predictions for the output, yet there are three main differences. First, the 
DBRNN provides a generic solution to the SISO problem without being restricted 
to dependency parsing. The target sequences in training (namely, $\underline{\boldsymbol{\hat{Y}}}^f_t$, 
$\underline{\boldsymbol{\hat{Y}}}^b_t$ and $\underline{\boldsymbol{\hat{Y}}}_t$) are the same for the DBRNN while the solution in 
\cite{BiDecoderParsing} has different target sequences. Second, the attention 
mechanism is used in \cite{BiDecoderParsing} but not in the DBRNN.

\section{Experiments}\label{sec:experiment}

\subsection{Experimental Setup}\label{sec:experiment_settings}

In the experiments, we compare the performance of five RNN macro-models: 
\begin{enumerate}
\item basic one-directional RNN (basic RNN);
\item bidirectional RNN (BRNN);
\item sequence-to-sequence (seq2seq) RNN \cite{Seq2Seq} (a variant of the encoder-decoder);
\item seq2seq with attention \cite{Grammar};
\item dependent bidirectional RNN (DBRNN), which is proposed in this work.
\end{enumerate}
For each RNN model, we compare three cell designs: LSTM, GRU, and ELSTM. 

We conduct experiments on three problems: part of speech (POS) tagging, language modeling \footnote{It 
asks the machine to predict the next word given all preceding words in a
sentence. It is also known as the automatic sentence generation task.} and
dependency parsing (DP). We report the testing accuracy for the POS tagging 
problem, the perplexity (i.e. the natural exponential of the model's cross-entropy loss) for LM and the unlabeled attachment score (UAS) and the labeled attachment 
score (LAS) for the DP problem. The POS tagging task is an easy one which
requires shorter memory while the LM and DP task demand longer memory.
For the latter two tasks, there exist more complex relations between the input and
the output. For the DP problem, we compare our solution with the
GRU-based bi-attention model (bi-Att). Furthermore, we compare the DBRNN 
using the ELSTM cell with two other non-RNN-based neural network
methods. One is transition-based DP with neural network (TDP) proposed
by Chen {\em et al.} \cite{NN-TransitionDP}. The other is convolutional
seq2seq (ConvSeq2seq) proposed by Gehring {\em et al.}
\cite{ConvSeq2seq}.  For the proposed DBRNN, we show the result for the final 
combined output (namely, $p_t$).  We adopt
$T_s=1$ in the basic RNN, BRNN, and DBRNN models and $T_s=100$ in the 
other two seq2seq models for the POS tagging problem. We use $T_s = 3$ and $T_s=100$ in 
all models for the LM problem and the DP problem, respectively.

The training, validation and testing dataset used for LM is from the Penn Treebank (PTB) \cite{PTB}. The PTB has 42,068, 3,370 and 3,761 training, validation and testing sentences respectively. It has in total 10,000 tokens. The training dataset used for the POS tagging and DP problems are from the Universal
Dependency 2.0 English branch (UD-English). It contains 12,543 sentences
and 14,985 unique tokens. The test dataset in both experiments is from
the test English branch (gold, en.conllu) of CoNLL 2017 shared task
development and test data.  The input to the POS tagging and the DP
problems are the stemmed and lemmatized sequences (column 3 in CoNLL-U
format). The target sequence for the POS tagging is the universal POS
tag (column 4). The target sequence for the DP is the interleaved
dependency relation to the headword (relation, column 8) and its
headword position (column 7). As a result, the length of the target
sequence for the DP is twice of the length of the input sequence. 

\begin{table}[htbp]
\centering
\caption{Training dataset.}\label{tab:dataset}
\begin{tabular*}{0.8\textwidth}{l c c c c} \hline
& \# Training & \# Validation & \# Testing & \# Tokens \\ \hline
PTB & 42,068 & 3,370 & 3,761 & 10,000 \\
UD 2.0 & 12,543 & 2,002 & 2,077 & 14,985\\
\end{tabular*}
\end{table}

\begin{table}[htbp]
\centering
\caption{Network parameters and training details.}\label{tab:setup}
\begin{tabular*}{0.7\textwidth}{l|c|c} \hline
Parameter & POS \& DP & LM \\ \hline
Embedding layer vector size& 512 & 5\\ \hline
Number of RNN cells & 512 & 5 \\ \hline
Batch size & 20 & 50 \\ \hline
Number of RNN layers& \multicolumn{2}{c}{1} \\\hline
Training steps & \multicolumn{2}{c}{11 epochs}\\ \hline
Learning rate & \multicolumn{2}{c}{0.5} \\ \hline
Optimizer & \multicolumn{2}{c}{AdaGrad\cite{AdaGrad}} \\ 
\end{tabular*}
\end{table}

The input is first fed into a trainable embedding layer \cite{Sparsity}
before it is sent to the actual network. Table \ref{tab:setup} shows the
detailed network and training specifications.  We do not finetune
network hyper-parameters or apply any engineering trick (e.g. feeding
additional inputs other than the raw embedded input sequences) for the
best possible performance since our main goal is to compare the
performance of the LSTM, GRU, ELSTM cells under various
macro-models. 

\begin{table}[htb]
\centering
\caption{LM test perplexity}\label{tab:LM_ppx}
\begin{tabular*}{0.8\textwidth}{@{\extracolsep{\fill} }l  c c c }
\hline
 & LSTM & GRU & ELSTM  \\ \hline 
BASIC RNN & 267.47 & 262.39 & \textbf{248.60}  \\ 
BRNN &  78.56 & 82.83 & \textbf{71.65} \\ 
Seq2seq  &  296.92 & 293.99 & \textbf{266.98} \\ 
Seq2seq with Att &  17.86 & 232.20 & \textbf{11.43} \\
DBRNN & 9.80 & 24.10 & \textbf{6.18}
\end{tabular*}
\end{table}

\begin{table}[htbp]
\centering
\caption{DP test results (UAS/LAS \%)  }\label{tab:DP_accu_ytm1}
\begin{tabular*}{\textwidth}{@{\extracolsep{\fill} }l c c c}
\hline
 & LSTM & GRU & ELSTM \\ \hline 
BASIC RNN & 43.24/25.28 & 45.24/29.92 & \textbf{58.49}/\textbf{36.10} \\ 
BRNN  & 37.88/25.26 & 16.86/8.95 & \textbf{55.97}/\textbf{35.13} \\ 
Seq2seq  & 29.38/6.05 & 36.47/13.44 & \textbf{48.58}/\textbf{24.05} \\ 
Seq2seq with Att & 31.82/16.16 & 43.63/33.98 & \textbf{64.30}/\textbf{52.60} \\
DBRNN  & 51.38/39.71 & 52.23/37.25 & \textbf{61.35}/\textbf{43.32} \\
Bi-Att  \cite{BiDecoderParsing} \tablefootnote{The result is generated by 
using exactly the same settings in Table. \ref{tab:setup}. 
We do not feed in the network with information other than input sequence 
itself.} & & 59.97/44.94 & \\
\end{tabular*}
\end{table}

\begin{table}[htb]
\centering
\caption{POS tagging test accuracy
(\%)}\label{tab:POS_accu_ytm1}
\begin{tabular*}{0.8\textwidth}{@{\extracolsep{\fill} }l  c c c }
\hline
 & LSTM & GRU & ELSTM  \\ \hline 
BASIC RNN & 87.30 & \textbf{87.51} & 87.44  \\ 
BRNN & \textbf{89.55} & 89.39 & 89.29  \\ 
Seq2seq  & 24.43 & 35.27 & \textbf{50.42}  \\ 
Seq2seq with Att & 31.34 & 34.60 & \textbf{81.72}  \\
DBRNN & \textbf{89.86} & 89.06 & 89.28 
\end{tabular*}
\end{table}

\begin{figure}[htbp]
\centering
\includegraphics[width=\linewidth]{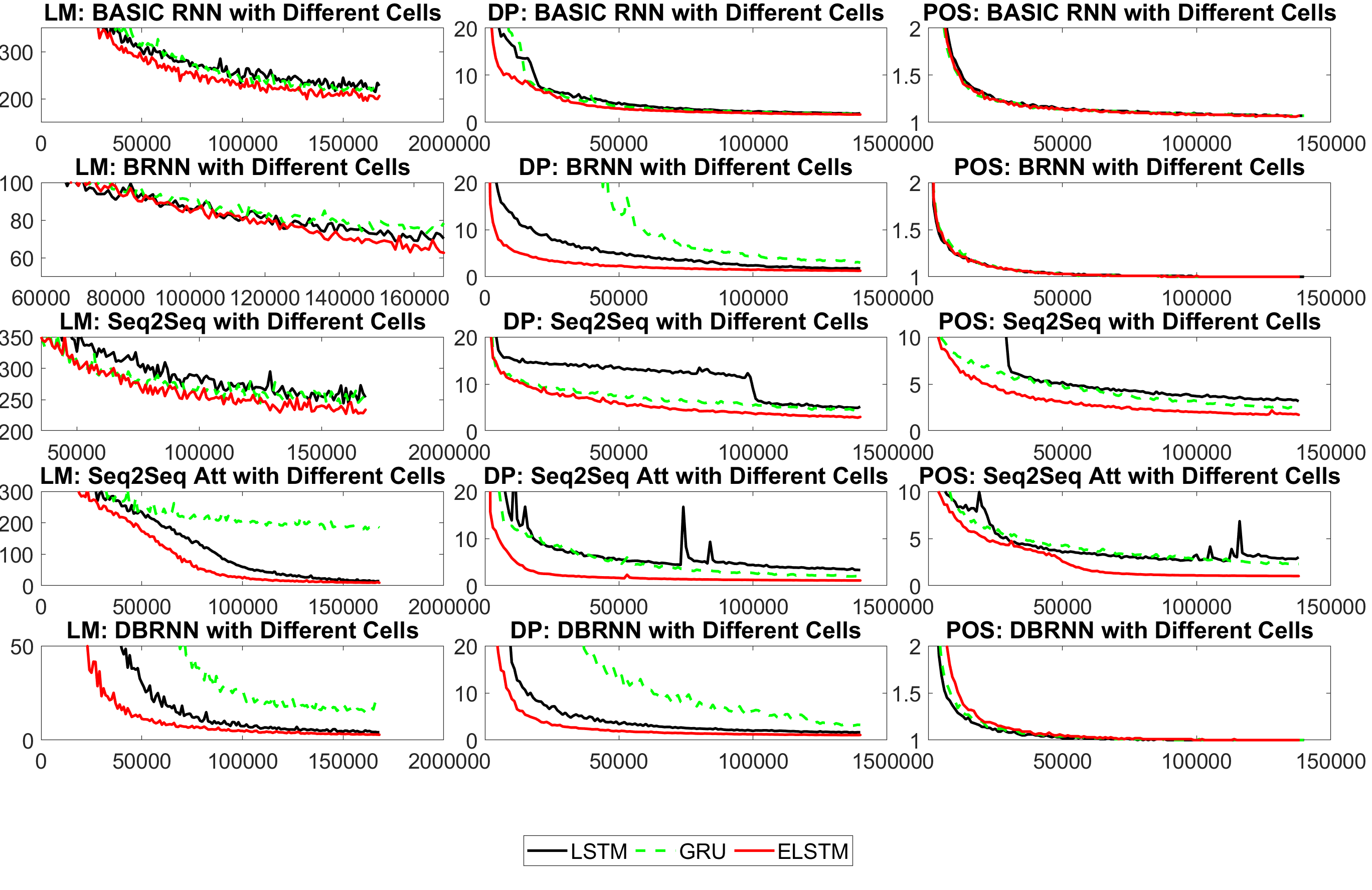}
\caption{The training perplexity vs. training steps of different cells.}\label{fig:cells_ppx}
\end{figure}

\begin{figure}[htbp]
\centering
\includegraphics[width=\linewidth]{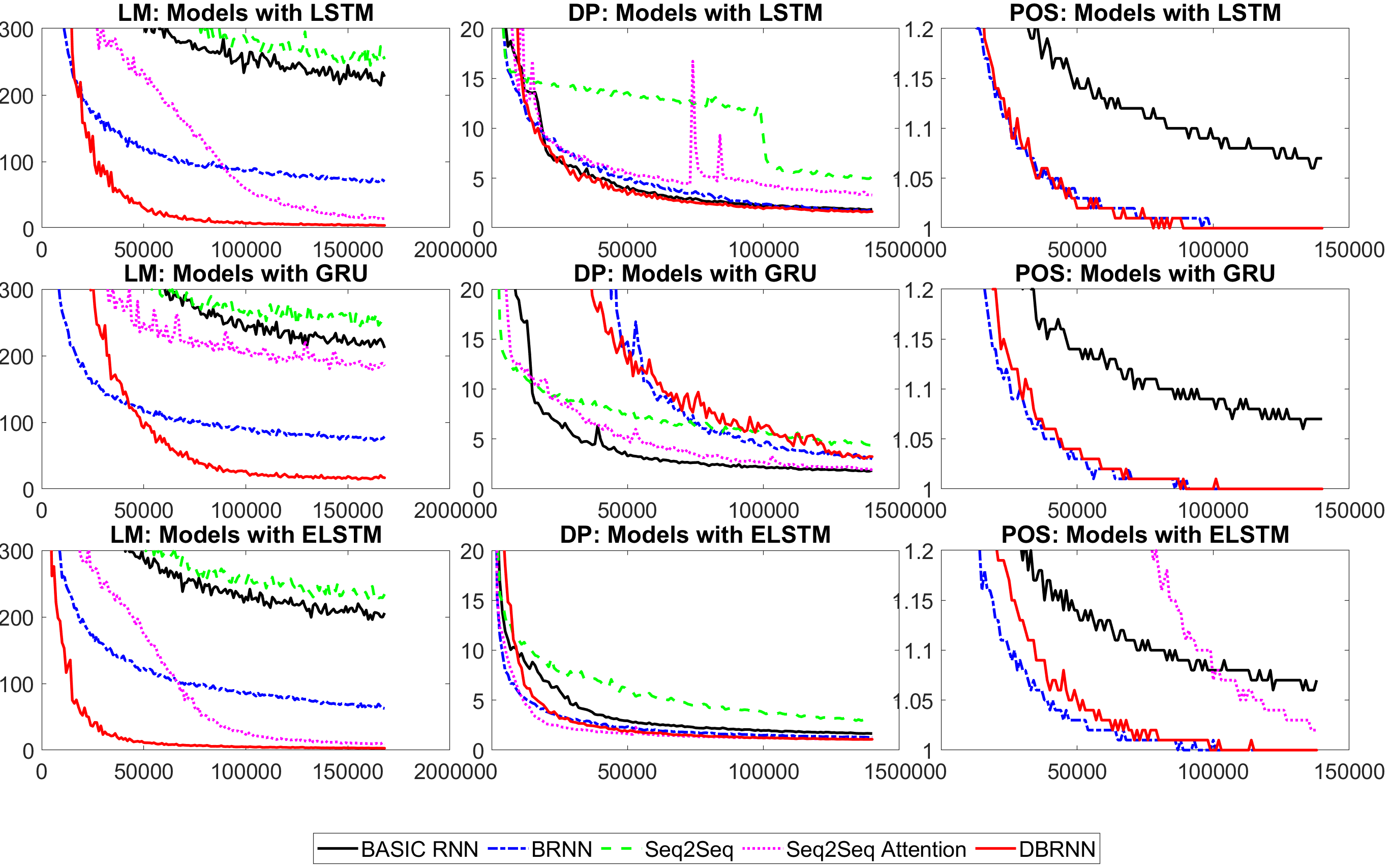}
\caption{The training perplexity vs. training steps of different macro models.}\label{fig:models_ppx}
\end{figure}

\subsection{Comparison of RNN Models}\label{sec:experiment_results_RNN}

The results of the LM, the DP and the POS tagging are shown in Tables
\ref{tab:LM_ppx} - \ref{tab:DP_accu_ytm1}, respectively. 

The training perplexity of different cell models and the macro models are shown in Fig. \ref{fig:cells_ppx} and \ref{fig:models_ppx} respectively. We see that the proposed ELSTM cell outperforms the LSTM and GRU
cells in most RNN models. This is especially true for complex language
tasks like LM and DP, where the ELSTM cell outperforms other cell designs by
a significant margin. The ELSTM cell even outperforms the bi-Att model, which was designed 
specifically for the DP task. This demonstrates the effectiveness of the
sequence of scaling factors adopted by the ELSTM cell.  It allows the
network to retain longer memory with better attention. For the simple POS tagging task, ELSTM also shows equal or better performance in comparison to other cell models. Overall, ELSTM delivers good performance across tasks with different complexity.

The ELSTM cell with large $T_s$ value perform
particularly well for the seq2seq (with and without attention) model.
The hidden state, $c_t$, of ELSTM cell is more expressive in
representing patterns over a longer distance. Since the seq2seq design
relies on the expressive power of a hidden state, ELSTM has a clear
advantage. 

Generally speaking, the scaling factor number depends on the memory length required by the specific task. For example, the output of POS tagging task is mostly a function of its immediate before, current and after inputs. Thus, the observation of using only one scaling factor gave the optimal performance for POS tagging makes sense since the memory length required for this task is mostly 1. The same observation goes for the language modeling task, where the memory length required is widely believed to be 3-5 and the best performance was achieved when the scaling factor number is 3. On the other hand, the encoder-decoder RNN behaves differently, where the usage of more than the required scaling factor number tends to yield better performance. This may have something to do with the memory length required at the decoder side is not the same as the one for the encoder. The underlining mechanism demands future study. Our recommendation is to estimate the required memory length before hyperparameter selection. If the estimation is not possible or reliable, one can begin by setting the scaling factor number to the maximum training sequence length and searching for its optimal value by binary search.

For the DBRNN, we see that the it achieves the best performance
across different macro models for the LM problem. It also outperforms
the BRNN and the seq2seq in both the POS tagging and the DP problems
regardless of the cell types. This shows its robustness. The DBRNN may
overfit to the training data in other cases. One may use a proper
regularization scheme in the training process to address it, which will
be an interesting future work item.

\begin{figure}[htbp]
\centering
\includegraphics[width=\linewidth]{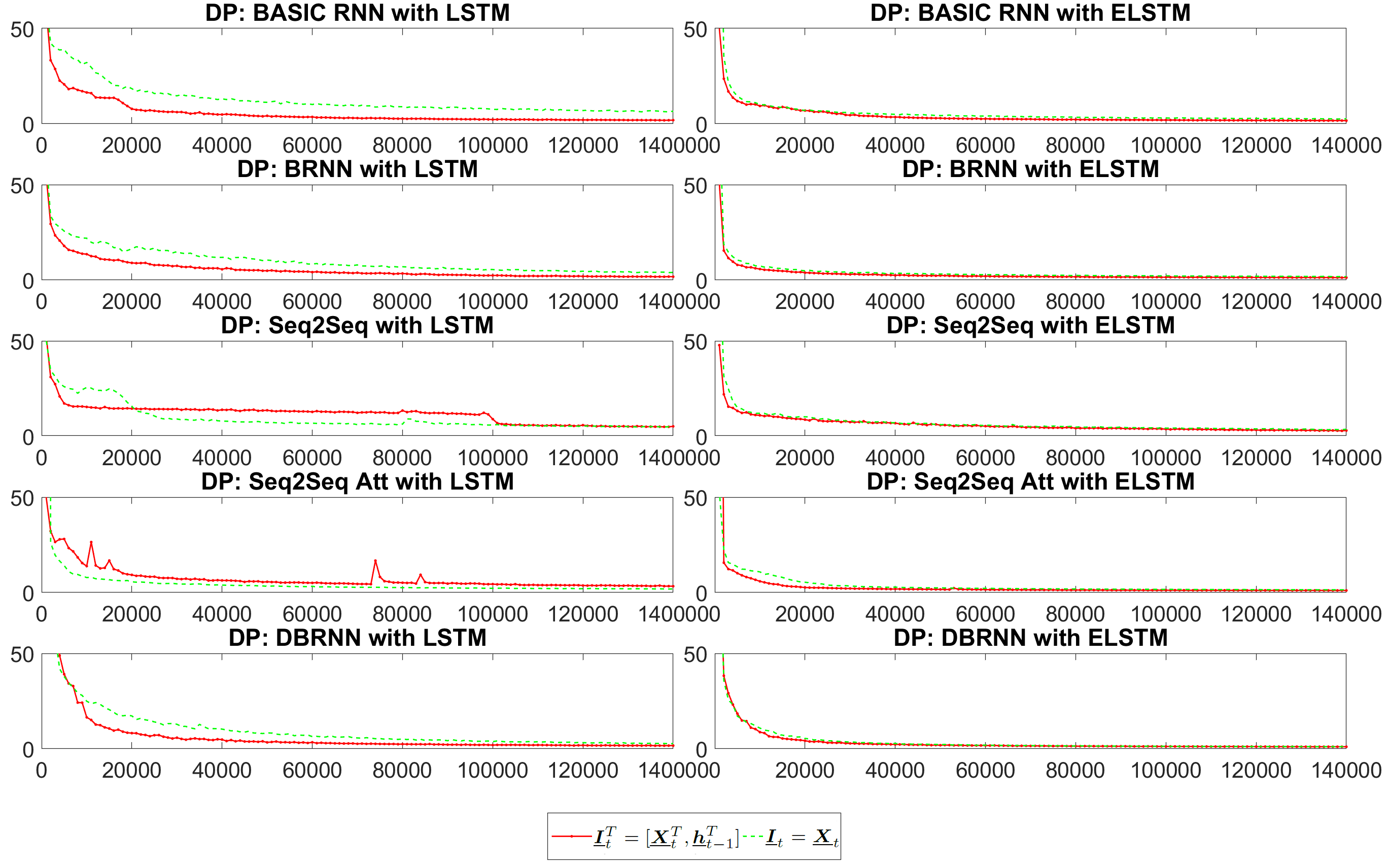}
\caption{The training perplexity vs. training steps of different models with $\underline{\boldsymbol{I}}_t = \underline{\boldsymbol{X}}_t$ and $\underline{\boldsymbol{I}}^T_t=[\underline{\boldsymbol{X}}^T_t, \underline{\boldsymbol{h}}^T_{t-1}]$ for the DP task.}\label{fig:prenopre_DP}
\end{figure}

To substantiate our claim in Sec. \ref{sec:memory}, we conduct
additional experiments to show the robustness of the ELSTM cell and the
DBRNN. Specifically, we compare the performance of the same five models
with LSTM, and ELSTM with $I_t = X_t$ for the same language
tasks. We do not include the GRU cell since it inherently demands $\underline{\boldsymbol{I}}_t^T
= [\underline{\boldsymbol{X}}_t^T, \underline{\boldsymbol{h}}_{t-1}^T]$. The convergence behaviors of $\underline{\boldsymbol{I}}_t = \underline{\boldsymbol{X}}_t$ and
$\underline{\boldsymbol{I}}_t^T = [\underline{\boldsymbol{X}}_t^T, \underline{\boldsymbol{h}}_{t-1}^T]$ with the LSTM, ELSTM cell
for the DP problem are shown in Fig. \ref{fig:prenopre_DP}. We
see that the ELSTM does not behave much differently
between $\underline{\boldsymbol{I}}_t = \underline{\boldsymbol{X}}_t$ and $\underline{\boldsymbol{I}}_t^T = [\underline{\boldsymbol{X}}_t^T, \underline{\boldsymbol{h}}_{t-1}^T]$ while the LSTM
does.  This shows the effectiveness of the ELSTM
design regardless of the input. More performance comparison will be
provided in the Appendix. 

\subsection{Comparison between ELSTM and Non-RNN-based Methods}\label{sec:experiment_comparison}

As stated earlier, the ELSTM design is more capable of extending the
memory and capturing complex SISO relationships than other RNN cells.
In this subsection, we compare the DP performance of two models built
upon the ELSTM cell (namely, the DBRNN and the seq2seq with attention)
and two non-RNN-based neural network based methods (i.e., the TDP
\cite{NN-TransitionDP} and the convseq2seq \cite{ConvSeq2seq}).  The TDP
is a hand-crafted method based on a parsing tree, and its neural network
is a multi-layer perceptron with one hidden layer. Its neural network is
used to predict the transition from a tail word to its headword. The
convseq2seq is an end-to-end CNN with an
attention mechanism.  We used the default settings for the TDP and the
convseq2seq as reported in \cite{NN-TransitionDP} and
\cite{ConvSeq2seq}, respectively. For the TDP, we do not use the ground
truth POS tags but the predicted dependency relation labels as the input
to the parsing tree for the next prediction. 

\begin{table}[htbp]
\centering
\caption{DP test accuracy (\%) and system settings}\label{tab:DP_nonRNN}
\begin{tabular*}{\textwidth}{@{\extracolsep{\fill} }l  c c c c }
\hline
 & Seq2seq-E  & DBRNN-E& Convseq2seq & TDP  \\ \hline 
UAS & \textbf{64.30} & 61.35  & 52.55 & 62.29 \\
LAS & \textbf{52.60} & 43.32 & 44.19 & 52.18 \\
Training steps & 11 epochs & 11 epochs & 11 epochs & 11 epochs \\
\# parameters & 12,684,468 & 16,460,468 &  22,547,124 &950,555 \\
Pretrained embedding & No & No & No & Yes \\
End-to-end & Yes & Yes & Yes & No\\
Regularization & No & No & No & Yes \\
Dropout & No & No & Yes & Yes \\
Optimizer & AdaGrad & AdaGrad & NAG \cite{NAG} & AdaGrad \\
Learning rate & 0.5 & 0.5 & 0.25 & 0.01 \\
Embedding size & 512 & 512 & 512 & 50 \\
Encoder layers & 1 & N/A & 4 & N/A \\
Decoder layers & 1 & N/A & 4 & N/A \\
Kernel size & N/A & N/A & 3 & N/A\\
Hidden layer size & N/A & N/A & N/A & 200 \\
\end{tabular*}
\end{table}

We see from Table \ref{tab:DP_nonRNN} that the ELSTM-based models learn
much faster than the CNN-based convseq2seq model with fewer parameters.
The convseq2seq uses dropout while the ELSTM-based models do not.
It is also observed that convseq2seq does not converge if Adagrad is used as 
its optimizer. The ELSTM-based seq2seq with attention even outperforms the
TDP, which was specifically designed for the DP task. Without a good
pretrained word embedding scheme, the UAS and LAS of TDP drop drastically to 
merely $8.93\%$ and $0.30\%$ respecively. 

\section{Conclusion and Future Work}\label{sec:conclusion}

Although the memory of the LSTM and GRU celles fades slower than that of
the SRN, it is still not long enough for complicated language tasks such
as dependency parsing. To address this issue, we proposed the ELSTM 
to enhance the memory capability of an RNN cell.
Besides, we presented a new DBRNN model that has the merits of both the
BRNN and the encoder-decoder.  It was shown by experimental results that
the ELSTM outperforms other RNN cell designs by a
significant margin for complex language tasks. The DBRNN model is
superior to the BRNN and the seq2seq models for simple and complex
language tasks. Furthermore, the ELSTM-based RNN models outperform the
CNN-based convseq2seq model and the handcrafted TDP. There are
interesting issues to be explored furthermore.  For example, is the
ELSTM cell also helpful in more sophisticated RNN models such as the
deep RNN?  Is it possible to make the DBRNN deeper and better?  They are
left for future study. 

\section{Declarations of interest}\label{sec:declarations}
Declarations of interest: none

\section{Acknowledgements}\label{sec:acknowledgments}
This research did not receive any specific grant from funding agencies in 
the public, commercial, or not-for-profit sectors.

\section*{References}

\bibliography{mybibfile}

\begin{thebibliography}{10}
\expandafter\ifx\csname url\endcsname\relax
  \def\url#1{\texttt{#1}}\fi
\expandafter\ifx\csname urlprefix\endcsname\relax\def\urlprefix{URL }\fi
\expandafter\ifx\csname href\endcsname\relax
  \def\href#1#2{#2} \def\path#1{#1}\fi

\bibitem{BiDecoderParsing}
H.~Cheng, H.~Fang, X.~He, J.~Gao, L.~Deng, Bi-directional attention with
  agreement for dependency parsing, In Proceedings of The Empirical Methods in
  Natural Language Processing (EMNLP 2016).

\bibitem{ConvSeq2seq}
J.~Gehring, G.~Auli~M, D.~Yarats, Y.~Denis, D.~Yann~N., Convolutional sequence
  to sequence learning, in: arXiv preprint, no. 1705.03122, 2017.

\bibitem{Time}
J.~Elman, Finding structure in time, Cognitive Science 14 (1990) 179--211.

\bibitem{Jordan}
M.~Jordan, Serial order: A parallel distributed processing approach, Advances
  in Psychology 121 (1997) 471--495.

\bibitem{LSTM}
S.~Hochreiter, J.~Schmidhuber, Long short-term memory, Neural Computation 9
  (1997) 1735--1780.

\bibitem{GRU}
K.~Cho, B.~v. Merrienboer, C.~Gulcehre, D.~Bahdanau, F.~Bougares, H.~Schwenk,
  Y.~Bengio, Learning phrase representations using {RNN} encoder–decoder for
  statistical machine translation, In Proceedings of The Empirical Methods in
  Natural Language Processing (EMNLP 2014).

\bibitem{BRNN}
M.~Schuster, K.~K. Paliwal, Bidirectional recurrent neural networks, Signal
  Processing 45 (1997) 2673--2681.

\bibitem{Seq2Seq}
I.~Sutskever, O.~Vinyals, Q.~V. Le, Sequence to sequence learning with neural
  networks, Advances in Neural Information Processing Systems (2014)
  3104--3112.

\bibitem{Grammar}
O.~Vinyals, L.~Kaiser, T.~Koo, S.~Petrov, I.~Sutskever, G.~Hinton, Grammar as a
  foreign language, Advances in Neural Information Processing Systems (2015)
  2773--2781.

\bibitem{BiEnc_Dec}
D.~Bahdanau, K.~Cho, Y.~Bengio, Neural machine translation by jointly learning
  to align and translate, In Proceedings of the International Conference on
  Learning Representations (ICLR 2015).

\bibitem{Deep_RNN}
R.~Pascanu, C.~Gulcehre, K.~Cho, Y.~Bengio, How to construct deep recurrent
  neural networks, arXiv:1312.6026.

\bibitem{OnDifficult}
P.~Razvan, T.~Mikolov, Y.~Bengio, On the difficulty of training recurrent
  neural networks, In Proceedings of The International Conference on Machine
  Learning (ICML 2013) (2013) 1310--1318.

\bibitem{Difficult}
Y.~Bengio, P.~Simard, P.~Frasoni, Learning long-term dependencies with gradient
  descent is difficult, Neural Networks 5 (1994) 157--166.

\bibitem{forget}
F.~A. Gers, J.~Schmidhuber, F.~Cummins, Learning to forget: Continual
  prediction with lstm, Neural Computation (2000) 2451--2471.

\bibitem{LSTM_Odyssey}
K.~Greff, R.~K. Srivastava, J.~Koutnik, B.~R. Steunebrink, J.~Schmidhuber,
  Lstm: A search space odyssey, IEEE transactions on neural networks and
  learning systems 28~(10) (2017) 2222--2232.

\bibitem{GRU_Empirical}
J.~Chung, C.~Gulcehre, K.~Cho, Y.~Bengio, Empirical evaluation of gated
  recurrent neural networks on sequence modeling, arXiv
  preprint~(arXiv:1412.3555).

\bibitem{Visualize_RNN}
A.~Karpathy, J.~Johnson, L.~Fei-Fei, Visualizing and understanding recurrent
  networks, arXiv preprint~(arXiv:1506.02078v2).

\bibitem{SENet}
J.~Hu, L.~Shen, G.~Sun, Squeeze-and-excitation networks, in: Proceedings of the
  IEEE conference on computer vision and pattern recognition (CVPR 2018), 2018,
  pp. 7132--7141.

\bibitem{NN-TransitionDP}
D.~Chen, M.~Christopher, A fast and accurate dependency parser using neural
  networks, in: In Proceedings of The Empirical Methods in Natural Language
  Processing (EMNLP 2014), 2014, pp. 740--750.

\bibitem{PTB}
M.~Marcus, B.~Santorini, M.~A. Marcinkiewicz, Building a large annotated corpus
  of english: the penn treebank, Computational Linguistics 19 (1993) 313--330.

\bibitem{AdaGrad}
Duchi, Adaptive subgradient methods for online learning and stochastic
  optimization, The Journal of Machine Learning Research (2011) 2121--2159.

\bibitem{Sparsity}
Y.~Bengio, R.~Ducharme, P.~Vincent, C.~Jauvin, A neural probabilistic language
  model, Journal of Machine Learning Research (2003) 1137--1155.

\bibitem{NAG}
Y.~Nesterov, A method of solving a convex programming problem with convergence
  rate o (1/k2), in: Soviet Mathematics Doklady, Vol.~27, 1983, pp. 372--376.

\end{thebibliography}
\bibliographystyle{elsarticle-num}

\newpage
\appendix

\section*{Appendix A: More Experimental Results}\label{app_experiment}

\begin{figure}[htbp]
\centering
\includegraphics[width=\linewidth]{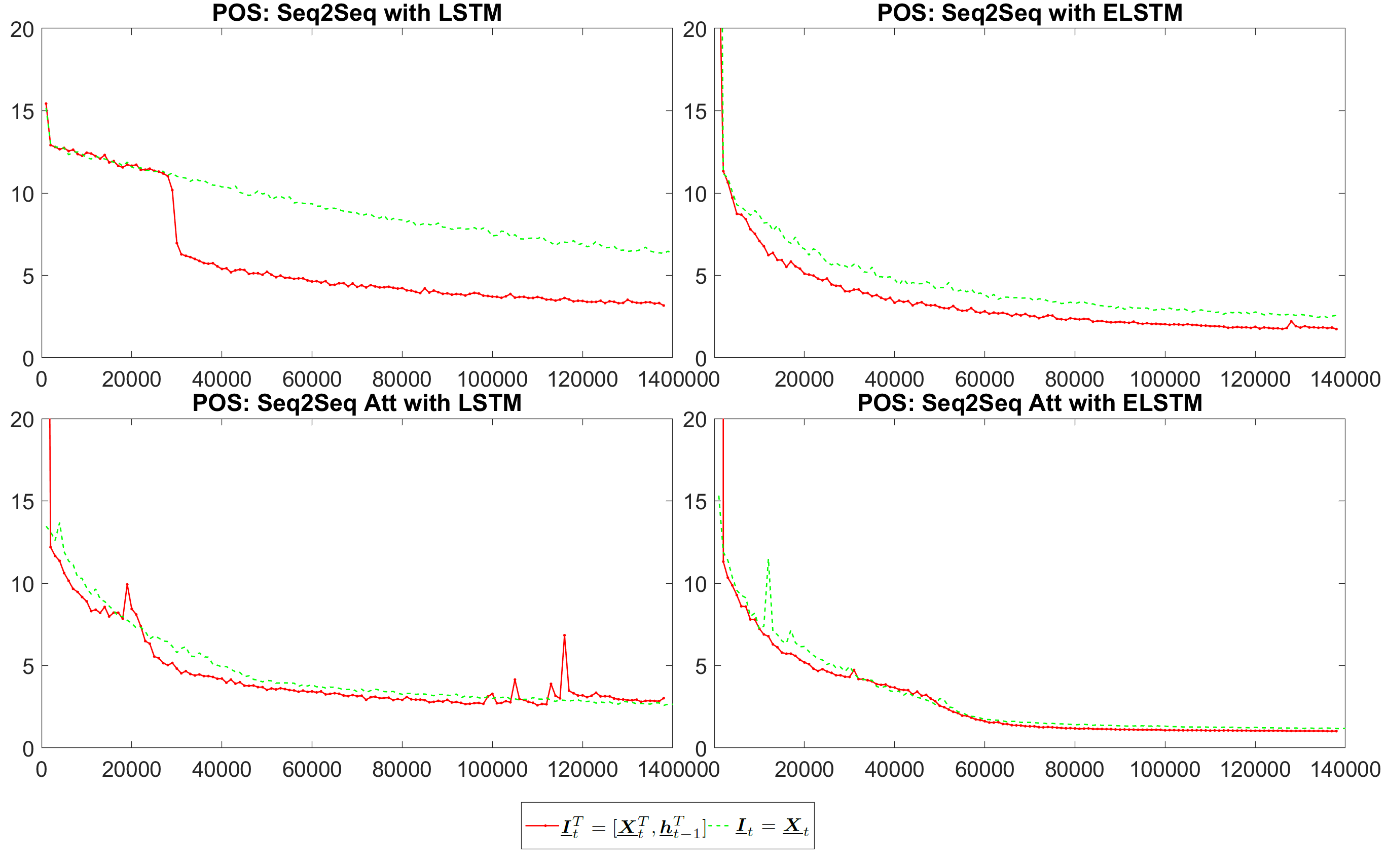}
\caption{The training perplexity vs. training steps of different models with $\underline{\boldsymbol{I}}_t = \underline{\boldsymbol{X}}_t$ and $\underline{\boldsymbol{I}}^T_t=[\underline{\boldsymbol{X}}^T_t, \underline{\boldsymbol{h}}^T_{t-1}]$ for the POS tagging task.}\label{fig:prenopre_POS}
\end{figure}

In the appendix, we compare
the training perplexity between $\underline{\boldsymbol{I}}_t = \underline{\boldsymbol{X}}_t$ and $\underline{\boldsymbol{I}}_t^T = [\underline{\boldsymbol{X}}_t^T,
\underline{\boldsymbol{h}}_{t-1}^T]$ for various models with the LSTM, and the ELSTM cells in Fig.
\ref{fig:prenopre_POS}.

\end{document}